\pdfoutput=1
\documentclass[11pt]{article}

\usepackage{acl}

\usepackage{times}
\usepackage{latexsym}

\usepackage[T1]{fontenc}
\usepackage[utf8]{inputenc}

\usepackage{microtype}

\usepackage{inconsolata}
\usepackage{dsfont}
\usepackage{hyperref}
\usepackage{booktabs}
\usepackage{tabularx}
\usepackage{multirow}
\usepackage{graphicx}
\usepackage{xcolor} % 支持表格内颜色设置
\usepackage{colortbl}
\usepackage{pifont}        % 勾叉符号
\usepackage{CJKutf8}    % 支持Unicode字符 ├ └ ─
\usepackage{tikz}
\usetikzlibrary{shapes.geometric, arrows, positioning}

\title{Probing Language Models for Pre-training Data Detection}
\author{
 \textbf{Zhenhua Liu}\textsuperscript{1}\: \textbf{Tong Zhu}\textsuperscript{1}\: \textbf{Chuanyuan Tan}\textsuperscript{1}\: \textbf{Haonan Lu}\textsuperscript{2}\: \textbf{Bing Liu}\textsuperscript{2}\: \textbf{Wenliang Chen}\textsuperscript{1}\thanks{\enspace Corresponding author}
 \\
 \textsuperscript{1}Institute of Artificial Intelligence, School of Computer Science and Technology, \\ Soochow University, China
 \\
 \textsuperscript{2}OPPO AI Center, China
 \\
 \texttt{\{zhliu0106, tzhu7, cytan17726\}@stu.suda.edu.cn}, \texttt{wlchen@suda.edu.cn}
 \\
 \texttt{\{luhaonan, liubing2\}@oppo.com}
}

\begin{document}
\maketitle

\begin{abstract}

    Large Language Models (LLMs) have shown their impressive capabilities, while also raising concerns about the data contamination problems due to privacy issues and leakage of benchmark datasets in the pre-training phase. Therefore, it is vital to detect the contamination by checking whether an LLM has been pre-trained on the target texts. Recent studies focus on the generated texts and compute perplexities, which are superficial features and not reliable. In this study, we propose to utilize the probing technique for pre-training data detection by examining the model's internal activations. Our method is simple yet effective and leads to more trustworthy pre-training data detection. Additionally, we propose ArxivMIA, a new challenging benchmark comprising arxiv abstracts from Computer Science and Mathematics categories. Our experiments demonstrate that our method outperforms all the baselines, and achieves state-of-the-art performance on both WikiMIA and ArxivMIA, with additional experiments confirming its efficacy\footnote{Our code and dataset are available at \url{https://github.com/zhliu0106/probing-lm-data}}.

\end{abstract}

\section{Introduction}

\begin{figure*}
    \centering
    \includegraphics[width=\linewidth]{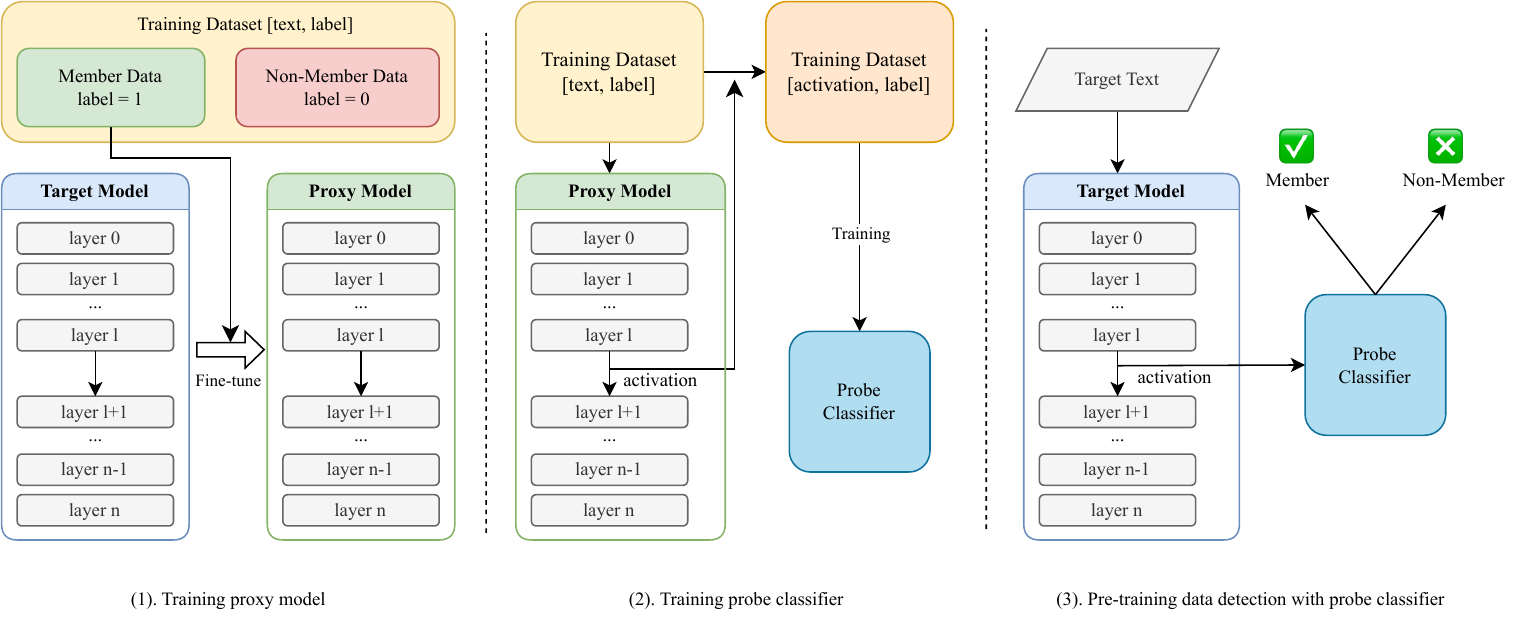}
    \caption{An overview of our method. Member data from the training dataset is first used to fine-tune the target model into a proxy model, from which activations are derived for training the probe classifier. The probe classifier then evaluates the target text to determine whether it was included in the model's pre-training data.}
    \label{fig:framework}
\end{figure*}

Large language models (LLMs) trained on massive corpora of texts demonstrate extraordinary abilities to understand, reason, and generate following natural language instructions \cite{brown2020language,anil2023palm}. Meanwhile, the open-source of LLMs has significantly contributed to the advancement and collaborative development within the LLM community \cite{zhang2022opt,touvron2023llama2,biderman2023pythia,qwen,2023internlm,llama-moe-2023}. Despite this progress, the lack of transparency raises ethical and legal questions, particularly about the use of potentially private data sourced from the internet, and threatens the reliability of benchmark evaluations due to the risk of leaking test data. Therefore, determining if certain texts have been utilized during the pre-training phase of the target LLM becomes a critical task.
%  in the pre-training of LLMs becomes a critical task.
% This step is vital for tackling these issues and ensuring the responsible development and deployment of LLM technologies.

Recent efforts to detect pre-training data in LLMs have attracted significant attention.
% Traditional approaches, such as n-gram based overlap comparison, require direct access to the pre-training datasets and have been critiqued for their high computational costs \cite{gao2020pile,brown2020language,dodge-etal-2021-documenting,chowdhery2023palm,anil2023palm,touvron2023llama1,touvron2023llama2}. 
Several studies have been proposed to investigate dataset contamination, including prompting LLMs to generate data-specific examples or using statistical methods to detect contamination in test sets \cite{sainz2023nlp,golchin2023data,oren2023proving}. Concurrently, Membership Inference Attacks (MIAs) in Natural Language Processing have been extensively explored for their potential to discern whether specific data was used in LLMs' pre-training \cite{carlini2021extracting,mireshghallah-etal-2022-quantifying,mattern-etal-2023-membership,shi2023detecting}.
% Despite their focus on evaluating superficial outputs of models, such as generated texts or loss metrics, to reach conclusions \cite{carlini2021extracting,mireshghallah-etal-2022-quantifying,mattern-etal-2023-membership,shi2023detecting}, these techniques mark a pivotal step towards understanding and mitigating privacy risks in LLM training processes.
The above solutions have achieved a certain success. However, they all rely on the model's superficial features, such as generated texts or loss metrics, which may not always be reliable indicators of pre-training data.
% we argue that using superficial features, like generated texts or loss metrics, is suboptimal since such information may not always be reliable.
% Despite making considerable progress in understanding and mitigating privacy risks in LLM pre-training processes, a common limitation is that they rely on the model's superficial outputs, such as generated texts or loss metrics, which may not always be reliable indicators of pre-training data .

Different from these conventional approaches, we propose a simple yet effective pre-training data detection method that utilizes the probing technique to examine the model's internal activations. This approach is based on the assumption: \textit{Texts that have been seen during the model's pre-training phase are represented differently in the model's internal activations compared to texts that have not been seen, so we could train a linear probe classifier to distinguish them.}
% In other words, the internal representations within a model contain encoded properties indicative of the data it has been trained on.

As illustrated in \autoref{fig:framework}, our method consists of three main steps: (1) We initiate our process by gathering a training dataset that the LLM has not previously been trained on, splitting the data into member and non-member subsets. We then inject the member data into the target model through a fine-tuning process on the member data alone. This step enables us to create a proxy model that retains the memory of the member data from the training dataset.
(2) Subsequently, we input both member and non-member subsets from the training dataset into the proxy model and extract the model's internal activations. These activations are employed to train a probe classifier that can distinguish between member and non-member data.
(3) Finally, given a target text, we can input it to the target model, extract the internal activations, and let the probe classifier infer whether the text is member data.
In other words, the probe classifier could assess whether the target text has been seen during the pre-training phase.

% the development and training of a proxy model, followed by the training of a probe classifier to scrutinize the model's internal activations. By fine-tuning the LLM with a subset of member data, we simulate the pre-training process, thereby creating a proxy model that mirrors the original model's behavior with respect to memorize pre-training data. The probe classifier is then trained on activations derived from this proxy model, enabling us to discern with greater accuracy whether a given sample was part of the pre-training corpus.

In order to evaluate various pre-training data detection methods in a more challenging scenario, we introduce ArxivMIA, a difficult benchmark in the academic domain. In contrast to the existing WikiMIA \cite{shi2023detecting} benchmark, which primarily utilizes event data from Wikipedia pages, ArxivMIA presents a more challenging scenario. The academic abstracts within ArxivMIA are rarer on the internet compared to Wikipedia content, naturally resulting in a lower duplication rate. Furthermore, the inherent complexity of texts targeted at researchers adds another layer of difficulty for model memorization. This combination of rarity and complexity makes it exceedingly challenging for large models to memorize such content during the pre-training process, making its detection through conventional methods markedly tougher. Therefore, ArxivMIA stands as an especially rigorous benchmark, highlighting the need for more sophisticated pre-training data detection methods.

% We conduct extensive experiments on both WikiMIA and ArxivMIA benchmarks and the results show that our approach not only surpassed all baseline methods but also achieved state-of-the-art performance, underscoring its superior performance and reliability. Furthermore, we conduct additional experiments to confirm the efficacy of our method. 
Our contributions can be summarized as follows:

\begin{itemize}
    \item[$\bullet$] We propose a novel pre-training data detection method that utilizes the probing technique. To the best of our knowledge, this is the first work to examine LLMs' internal activations to determine whether a given text was included in the pre-training data.
\end{itemize}

\begin{itemize}
    \item[$\bullet$] We propose ArxivMIA, a new benchmark in the academic domain to assess pre-training data detection methods.
          % under more challenging conditions. 
          % As an especially rigorous benchmark, 
          With a low duplication rate and the inherent complexity of texts, ArxivMIA presents a more challenging scenario and highlights the need for more sophisticated pre-training data detection methods.
\end{itemize}

\begin{itemize}
    \item[$\bullet$] We conduct extensive experiments on WikiMIA and ArxivMIA benchmarks. In addition, we also evaluate various detection methods on a downstream task datasets contamination challenge. Through comprehensive experimentation, we demonstrate that our proposed method outperforms all baselines, and achieves state-of-the-art performance.
\end{itemize}

\section{Related Work}

Related work involves membership inference attacks in NLP and dataset contamination.

\paragraph{Membership Inference Attacks in NLP.} Membership Inference Attacks (MIAs) are designed to identify if a particular data sample was included in the training dataset of a machine learning model \cite{shokri2017membership, yeom2018privacy,hu2022membership}. Most MIAs take a black-box setting, assuming that the adversary only has access to the model confidence or loss scores \cite{yeom2018privacy,sablayrolles2019white,jayaraman2021revisiting,watson2021importance}. Unlike it, similar to \citet{leino2020stolen}, we consider a white-box setting where the adversary has access to the model weights and activations. Specifically in NLP, a lot of studies have been proposed \cite{carlini2021extracting,mireshghallah-etal-2022-quantifying,mattern-etal-2023-membership,shi2023detecting}. \citet{carlini2021extracting} and \citet{mireshghallah-etal-2022-quantifying} separately investigated Likelihood Ratio Attacks for causal language models and masked language models. \citet{mattern-etal-2023-membership} proposed a neighbor attack that compares model loss for a given sample to losses of synthetically generated neighbor texts. \citet{shi2023detecting} measured the likelihood of outlier words in a given text, thereby assessing whether the text was likely part of a model's pre-training corpora. Similar to \citet{shi2023detecting}, we aim to detect pre-training data in LLMs. However, different from existing attacks that rely on the model's superficial features, we focus on the LLMs' internal activations, and the experiments show that our method outperforms existing attacks.

\paragraph{Dataset Contamination.} The dataset contamination in LLMs has been widely studied since benchmark datasets are intentionally or unintentionally included in pre-training corpora. The n-gram based overlap comparison methods not only require access to training corpora but take a long time to compute \cite{gao2020pile,brown2020language,dodge-etal-2021-documenting,chowdhery2023palm,anil2023palm,touvron2023llama1,touvron2023llama2}. Without access to pre-training corpora, there are also some methods to detect dataset contamination. \citet{sainz2023nlp} prompted LLMs to generate verbatim examples of a dataset split. \citet{golchin2023data} proposed the 'Data Contamination Quiz', which employs a multiple-choice format to assess a model's ability to recognize original dataset instances among perturbed versions. \citet{oren2023proving} presented a statistical test to demonstrate test set contamination in language models, leveraging the concept of exchangeability in benchmark datasets and comparing model log probabilities against shuffled dataset permutations.

\section{Methodology}

\subsection{Overview}
Different from conventional detection methods in MIA that rely on the model's superficial features, we directly analyze the model's internal activations, providing a deeper insight into its pre-training history. Our method employs the probe technique originally proposed by \citet{alain2016understanding}.
% This technique hypothesizes that the internal representations of a model inherently contain specific encoded properties, so we could train a linear probe classifier with logistic regression for identifying directions representing a feature. 
This technique hypothesizes that the internal representations of a model inherently contain specific encoded properties, so we could train a linear probe classifier with logistic regression for the detection of these properties.
In our context, we are interested in determining whether a text sample was included in the model's pre-training dataset. The framework of our method is illustrated in \autoref{fig:framework}.

\subsection{Task Definition}
The task of pre-training data detection follows a white-box setting of MIAs where the adversary has access to the model $\mathcal{M}$'s architecture and weights, but not the pre-training data.
% $\mathcal{D} _{pretrain}$
The adversary aims to determine whether a sample $s$ was included in the pre-training data
% $\mathcal{D} _{pretrain}$ 
of the model $\mathcal{M}$ with an attack method $A$: $ A_{\mathcal{M}}(s)\rightarrow \left\{0, 1\right\} $, where 1 represents member (seen) data, 0 denotes non-member (unseen) data. Usually, we have a scoring function $f$, then can get the confidence score $f(s) \in [0, 1]$, which represents the probability of the sample being a part of the pre-training dataset. Then we can classify the sample as a member or non-member based on a threshold $\gamma$ :
$$ A_{\mathcal{M}}(s)=\mathds{1}\left[f(s)< \gamma \right] $$

\subsection{Training Proxy Model}

Training a probe classifier needs a dataset $\left\{ \langle x_i, y_i \rangle \right\}$, where $x_i$ represents the sample's activation, and $y_i$ is a binary label indicating whether the sample is a member or non-member data. However, the absence of pre-training data makes it impossible to obtain the activations of the member or non-member samples. Therefore, we first gather a training dataset that the LLM has not previously been trained on, splitting the data into member and non-member subsets, and inject the member data into the proxy model to simulate data contamination, as detailed in \autoref{sec:training_dataset_collection}. The training dataset is $ D= \left\{ \langle s_i, y_i \rangle \right\}$, where $s_i$ represents the sample, and $y_i$ is a binary label indicating whether the sample is a member or non-member.

\paragraph{Prompt Template for Sample.} To standardize the input for consistency, each sample of $D$ is processed using a prompt template. In the subsequent experiments, we use the following prompt template: "\textit{Here is a statement: [SAMPLE] \textbackslash n Is the above statement correct? Answer:}".
% This prompt template is crucial for standardizing the input for consistency. 
Intuitively, we hope the last token of the sample can capture the information of it being member or non-member, and we believe the prompt template can help the target model to focus on this information. Further prompt template selection will be discussed in \autoref{sec:prompt_template_selection}.

\paragraph{Training Proxy Model.} Next we need to inject the member samples $ D_{member} = \left\{ s_i \mid \langle s_i, y_i \rangle \in D, y_i = 1 \right\}$ into the model $\mathcal{M}$ to let it memorize the member data. This injection is accomplished by fine-tuning the model on $D_{member}$. This step aims to make the model $\mathcal{M}$ memorize the member data, and the fine-tuning process is used to simulate the pre-training process. After this, we can get the proxy model $\mathcal{M}'$, which retains the memory of $D_{member}$. The proxy model $\mathcal{M}'$ is then used to generate the member and non-member sample activations $x$ for the dataset $D$.

\subsection{Training Probe Classifier}

The probe classifier takes the form $P_{\theta}(x) = \sigma(Wx)$, where $\sigma$ denotes the sigmoid function and $W$ represents the trainable weights. After obtaining the proxy model $\mathcal{M}'$, we construct the training dataset with $D$ for the probe.

In the paper, we focus on causal language models, which are trained to predict the next word given the previous words. In order to capture the representation of the sample, for each sample $\langle s_i, y_i \rangle$ in $D$, we extract the activation $x_l$ from the final token of the input in layer $l$ of the model $\mathcal{M}'$. Finally, we get the dataset $\{ \langle x^l_i, y_i \rangle \}$, which is used to train the probe $P_{\theta}$ with logistic regression.

\subsection{Pre-training Data Detection with Probe Classifier}

Given a benchmark, we already trained a probe $P_{\theta}$, which can be used to detect whether a sample is included in the pre-training data. For a sample $s$, we preprocess it with the above prompt template, then feed it into the model $\mathcal{M}$ to get the activation $x^l$. Finally, we can get the confidence score $P_{\theta}(x^l)$, which represents the probability of the sample being a part of the pre-training dataset. This score is then utilized to classify the sample as a member or non-member based on a threshold $\gamma$ :
$$ A_{\mathcal{M}}(s)=\mathds{1}\left[P_{\theta} (x^l)< \gamma \right] $$

\section{Data Construction}

\subsection{ArxivMIA}
\label{sec:arxivmia}

To evaluate various pre-training data detection methods in a more challenging scenario, we introduce ArxivMIA, a new benchmark comprising abstracts from the fields of Computer Science (CS) and Mathematics (Math) sourced from Arxiv. In contrast to the existing WikiMIA \cite{shi2023detecting} benchmark, which primarily utilizes event data from Wikipedia pages—thus prone to higher duplication rates in pre-training datasets—ArxivMIA presents a more challenging scenario.
% The academic abstracts within ArxivMIA are rarer on the internet compared to Wikipedia content, naturally resulting in a lower duplication rate. Furthermore, the inherent complexity of texts targeted at researchers adds another layer of difficulty for model memorization.

For dataset construction, abstracts published post-2024 are designated as non-member data, while member data are derived from the Arxiv subset of the RedPajama dataset \cite{together2023redpajama}. The RedPajama dataset is the reproduction of the LLaMA \cite{touvron2023llama1} training dataset and is extensively utilized in pre-training LLMs \cite{zhang2024tinyllama,openlm2023openllama}. This makes ArxivMIA particularly suited for testing LLMs pre-trained on the RedPajama dataset. Detailed information about ArxivMIA is presented in \autoref{tabs:datasets}. In summary, ArxivMIA has three distinctive features: Firstly, it is a larger dataset with a total of 2000 samples. Secondly, it covers multiple fields, including CS and Math. Lastly, it features a longer average sentence length, with an average of 143.1 tokens per sample. These characteristics make ArxivMIA a more challenging benchmark for evaluating pre-training data detection methodologies.

\begin{CJK}{UTF8}{gkai}
\begin{table*}[ht]
    \centering
    \begin{tabular}{lcccccc}
        \toprule
        \textbf{Dataset} & \textbf{Avg. Tokens} & \textbf{Members} & \textbf{Non-Members} & \textbf{Total} & \textbf{Real} & \textbf{Synthetic} \\
        % \midrule
        \cmidrule(lr){1-1} \cmidrule(lr){2-5} \cmidrule(lr){6-7}
        WikiMIA          & 32.0                 & 387              & 289                  & 676            & 100*          & 100                \\
        % \midrule
        ArxivMIA         & 143.1                & 1,000             & 1,000                 & 2,000           & 200           & 200                \\
        % \cmidrule{2-7}
        \quad ├ ArxivMIA-CS      & 181.8                & 400              & 400                  & 800            & 80            & 80                 \\
        \quad └ ArxivMIA-Math    & 117.2                & 600              & 600                  & 1,200           & 120           & 120                \\
        \bottomrule
    \end{tabular}
    \caption{Information of Datasets. Real denote the number of collected real training data, and Synthetic denote the number of synthetic training data. * For convenience, we directly segregated a subset of 100 non-member data from WikiMIA as real data}
    \label{tabs:datasets}
\end{table*}
\end{CJK}

\subsection{Training Dataset Collection}
\label{sec:training_dataset_collection}

Our method needs to construct a training dataset $ D= \left\{ \langle s_i, y_i \rangle \right\}$ similar to the target benchmark, where $s_i$ represents the sample, and $y_i$ is a binary label indicating whether the sample is a member or non-member. This dataset is pivotal for training the probe to accurately evaluate the likelihood of a sample being included in the pre-training data. However, the construction of a training dataset for the probe is challenging due to the lack of access to the pre-training data. To address this, we propose a heuristic method:

\paragraph{Data Collection.} Firstly, we need to collect a dataset $D= \left\{ s_i \right\}$, and make sure they are not included in the pre-training data. There are two ways to accomplish it: (1). \textbf{Real data}: We collect the data published after the model release date. (2). \textbf{Synthetic data}: We can also use ChatGPT \cite{achiam2023gpt} to synthesize similar data according to the data to be detected. The detailed process is described in \autoref{sec:chatgpt_synthetic_data}.

\paragraph{Dataset Split.} Next, we randomly label half of the data in $D$ as non-member data, and the other half as member data. Then we get the dataset $D= \left\{\langle s_i, y_i \rangle \right\}$.

\vspace{6pt}

We constructed both real and synthetic training datasets for each benchmark respectively, with specifics outlined in \autoref{tabs:datasets}. Notably, for convenience, we directly segregated a subset of 100 non-member data from WikiMIA as real data, and the remaining part will be used in subsequent experiments.

\section{Experiments}
\label{sec:experiments}

We evaluate the performance of our method and other baselines against open-source language models trained to predict the next word, including Pythia-2.8B \cite{biderman2023pythia}, OPT-6.7B \cite{zhang2022opt}, TinyLLaMA-1.1B \cite{zhang2024tinyllama} and OpenLLaMA-13B \cite{openlm2023openllama}.

\subsection{Datasets}

\label{sec:datasets}

\paragraph{WikiMIA} proposed by \citet{shi2023detecting}, a dynamic benchmark designed to periodically and automatically evaluate detection methods on any newly released pre-trained LLMs. We use the WikiMIA-32 split in this work, which contains 776 samples with a max length of 32 tokens.

\paragraph{ArxivMIA} proposed in this work, a more complex benchmark comprising abstracts in the fields of Computer Science and Mathematics from Arxiv. The details refer to \autoref{sec:arxivmia}.

\vspace{6pt}

We split each dataset into a validation set and a test set in a ratio of 1:4. The validation set is used to select the best hyperparameters, and the test set is used to evaluate the performance of the detection methods.

\subsection{Evaluation Metrics}
Following \citet{shi2023detecting,carlini2022membership,mattern-etal-2023-membership}, we assess the effectiveness of detection methods using these metrics:

\paragraph{Area Under the ROC Curve (AUC).} The ROC curve plots the true positive rate (power) against the false positive rate (error) across various thresholds $\gamma$, which captures the trade-off between power and error. Therefore, the area under the ROC curve (AUC) serves as a singular, threshold-independent measure to evaluate the effectiveness of the detection method.

\paragraph{True Positive Rate (TPR) under low False Positive Rates (FPR).} We report TPR under low FPR by adjusting the threshold value $\gamma$. Concretely, we set 5\% as the target FPRs, and report the corresponding TPRs.

\subsection{Baselines}
To compare the performance of Probe Attack, we consider the following reference-free methods:

\paragraph{Loss Attack} proposed by \citet{yeom2018privacy}, which assesses the membership of the target sample based on the loss of the target model.

\paragraph{Neighbor Attack} proposed by \citet{mattern-etal-2023-membership}, which compares model loss for the target sample to losses of synthetically generated neighbor texts. We construct 100 neighbors for each sample using one-word replacement with the RoBERTa-base model \cite{liu2019roberta}.

\paragraph{Min-K\% Prob} proposed by \citet{shi2023detecting}, begins by calculating the probability of each token in the target sample, then selects the k\% of tokens with the lowest probabilities to compute their average log-likelihood. A high average log-likelihood suggests that the text is likely part of the pretraining data.

\vspace{6pt}

Following \citet{carlini2021extracting} and \citet{shi2023detecting}, we also consider reference-based methods, which calibrate difficulty by quantifying the intrinsic complexity of a target sample:

\paragraph{Comparing to Zlib Compression.} We compute the zlib entropy of the sample, which is the entropy in bits when the sequence is compressed using zlib\footnote{\url{https://github.com/madler/zlib}}. The detection score is then determined by the ratio of the model's perplexity to the zlib entropy.
% : the number of bits of entropy when the sequence is compressed with zlib compression. We then use the ratio of model perplexity and the zlib entropy as the detection score.

\paragraph{Comparing to Lowercased Text.} We compute the ratio of the perplexity of the sample before and after converting it to lowercase.
% This method is effective because lowercasing can significantly change the perplexity of memorized content that relies on specific casing.

\paragraph{Comparing to Smaller Model.} We compute the sample perplexity ratio of the target model to a smaller model pre-trained on the same data.

\subsection{Implementation Details}

For WikiMIA, we employ Pythia-2.8B \cite{biderman2023pythia} and OPT-6.7B \cite{zhang2022opt} as the target model following \citet{shi2023detecting}. For ArxivMIA, we employ TinyLLaMA-1.1B \cite{zhang2024tinyllama} and OpenLLaMA-13B\cite{openlm2023openllama} pre-trained on RedPajama \cite{together2023redpajama} as the target model.

For comparing to smaller model baseline setting, we take Pythia-70M for Pythia-2.8B, OPT-350M for OPT-6.7B, and OpenLLaMA-3B for OpenLLaMA-13B. Because there is no smaller model for TinyLLaMA, we leave this baseline out for TinyLLaMA.

For the training of the proxy model, we conducted a grid search hyperparameters on a held-out validation set in order to better inject member data into the model. Based on the performance, the best choice is to put all the data to be injected into one batch and train for 2 epochs. For different models and datasets, we set the best learning rate and activation extraction model layer according to the performance of the validation set. All experiments are conducted with 2 NVIDIA A100 (40GB) GPUs.

\section{Results and Analyses}

In this section, we report our main result and conduct ablation studies to analyze the impact of model size and training data number for our method. We also compare the performance of various detection methods on PubMedQA \cite{jin-etal-2019-pubmedqa} and CommonsenseQA \cite{talmor-etal-2019-commonsenseqa} in the contamination detection challenge proposed by \citet{oren2023proving}.

\subsection{Main Results}

\begin{table*}[h!]
    \centering
    \begin{tabular}{lcccccccc}
        \toprule
        \multirow{2}{*}{\textbf{Method}} & \multicolumn{2}{c}{\textbf{WikiMIA}} & \multicolumn{2}{c}{\textbf{ArxivMIA}} & \multicolumn{2}{c}{\textbf{ArxivMIA-CS}} & \multicolumn{2}{c}{\textbf{ArxivMIA-Math}} \\
                                & Pythia & OPT & TinyL. & OpenL. & TinyL. & OpenL.& TinyL. & OpenL.        \\
        \cmidrule(lr){1-1} \cmidrule(lr){2-3} \cmidrule(lr){4-5} \cmidrule(lr){6-7} \cmidrule(lr){8-9}
        \rowcolor[gray]{.93}\textbf{Reference-free Methods} & \multicolumn{8}{c}{} \\
        Loss Attack             & 63.9  & 63.0  & 45.1  & 49.1  & 45.3   & 51.4   & 44.9   & 47.4          \\
        Neighbor Attack         & 62.1  & 58.5  & 54.8  & 55.4  & 59.3   & 59.3   & 53.4   & 54.1          \\
        Min-K\% Prob            & 62.7  & 63.2  & 45.5  & 49.2  & 45.0   & 50.2   & 45.8   & 48.5          \\
        \midrule
        \rowcolor[gray]{.93}\textbf{Reference-based Methods} & \multicolumn{8}{c}{} \\
        Zlib Compression        & 63.8  & 62.9  & 42.9  & 43.8  & 38.0   & 40.4   & 44.0   & 44.7          \\
        Lowercased Text         & 64.7  & 61.6  & 46.8  & 50.2  & 43.8   & 47.8   & 48.4   & 50.8          \\
        Smaller Model           & 65.5  & 65.8  & -     & 55.9  & -   & 54.9   & -      & 56.7          \\
        \midrule
        \rowcolor[gray]{.93}\textbf{Our Method} & \multicolumn{8}{c}{} \\
        Probe w. Real Data      & \textbf{69.8}  & \textbf{68.1}  & 57.1  & 60.0  & 63.7   & 67.2   & 56.1   & 56.9          \\
        Probe w. Synthetic Data & 69.4  & 66.2  & \textbf{59.2}  & \textbf{60.3}  & \textbf{64.3}   & \textbf{67.3}   & \textbf{56.7}   & \textbf{57.4}          \\
        \bottomrule
    \end{tabular}
    \caption{AUC values of different methods on WikiMIA and ArxivMIA. TinyL. denotes TinyLLaMA, OpenL. denotes OpenLLaMA. We highlight the best results in \textbf{bold}.}
    \label{tabs:main_result}
\end{table*}

\begin{table*}[h!]
    \centering
    \setlength\tabcolsep{4pt} % 将列间距设置为4pt
    \begin{tabular}{lccccc}
        \toprule
        \multirow{2}{*}{\textbf{Method}} & \multicolumn{2}{c}{\textbf{WikiMIA}} & \multicolumn{2}{c}{\textbf{ArxivMIA}} & \multirow{2}{*}{\textbf{Avg.}} \\
                          & Pythia      & OPT  & TinyL. & OpenL. & \\
        \cmidrule(lr){1-1} \cmidrule(lr){2-3} \cmidrule(lr){4-5} \cmidrule(lr){6-6}
        \rowcolor[gray]{.93}\textbf{Reference-free Methods}  & \multicolumn{5}{c}{}                                                                    \\
        Loss Attack                      & 13.7                 & 11.4          & 5.1             & 5.6             &      9.0        \\
        Neighbor Attack                  & 14.0                 & 13.4          & 6.5             & 7.3             &      10.3        \\
        Min-K\% Prob                     & 16.9                 & 15.0          & 4.5             & 5.1             &      10.4        \\
        \midrule
        \rowcolor[gray]{.93}\textbf{Reference-based Methods} & \multicolumn{5}{c}{}                                                                    \\
        Zlib Compression                 & 17.3                 & 14.4          & 2.5             & 3.5             &       9.4       \\
        Lowercased Text                  & 10.1                 & 9.1           & 4.3             & 6.3             &       7.5       \\
        Smaller Model                    & 14.0                 & 10.5          & -               & \textbf{8.5}    &         11.0     \\
        \midrule
        \rowcolor[gray]{.93}\textbf{Our Method}              & \multicolumn{5}{c}{}                                                                    \\
        Probe w. Real Data               & 16.7                 & \textbf{15.4} & 7.5             & 7.4             &       \textbf{11.8}       \\
        Probe w. Synthetic Data          & \textbf{19.6}        & 10.5          & \textbf{8.6}    & 6.8             &       11.4       \\
        \bottomrule
    \end{tabular}
    \caption{True positive rates for different methods at 5\% positive rates on WikiMIA and ArxivMIA datasets. TinyL. denotes TinyLLaMA, OpenL. denotes OpenLLaMA. Best results are highlighted in \textbf{bold}.}
    \label{tabs:tpr_at_low_fpr}
\end{table*}

We present the main results of our experiments in \autoref{tabs:main_result} and \autoref{tabs:tpr_at_low_fpr}, where the former shows AUC values and the latter shows true positive rates at 5\% false positive rates. The results show that our method consistently outperforms all baselines on both WikiMIA and ArxivMIA benchmarks, achieving state-of-the-art AUC values. We also achieve the state-of-the-art average true positive rates at 5\% false positive rates across all detection methods on both benchmarks. We can further observe that:

\begin{itemize}
    \item The average performance across all detection methods is notably lower on ArxivMIA compared to WikiMIA. This disparity underscores the increased complexity of ArxivMIA as a benchmark. In addition to our method, the Neighbor Attack method exhibits a relatively competent AUC value.
    \item The performance gap between various detection methods across the two fields of ArxivMIA is notable. Specifically, in the ArxivMIA-CS, the average AUC value is comparatively higher, with our method achieving its best results above 60. In contrast, in the ArxivMIA-Math, the values are only above 50, differing by approximately 10 points. This discrepancy may suggest that mathematical content in academic papers is more challenging for Large Language Models (LLMs) to memorize, and consequently, harder for our method to detect.
    \item As shown in \autoref{tabs:main_result}, we separately test our method with real and synthetic data. On WikiMIA, the utilization of real data marginally outperforms synthetic data, while the opposite is observed on ArxivMIA. Despite a slight difference, the performance of our method is far superior to other baselines with both real and synthetic data.
\end{itemize}

\subsection{Cross-Domain Evaluation}
We conducted cross-domain evaluations with TinyLLaMA. As shown in \autoref{tabs:cross_domain}, we find that in-domain performance is better than cross-domain performance. Training on WikiMIA and then testing on ArxivMIA, AUC values dropped by 5 points. Conversely, training on ArxivMIA and then testing on WikiMIA, AUC values only decreased by 1 point. This supports the finding that ArxivMIA is relatively more challenging than WikiMIA.

\subsection{Ablation Studies}

We further investigate the impact of model size and training data number for our method:

\vspace{6pt}

\begin{figure}
    \centering
    \includegraphics[width=\linewidth]{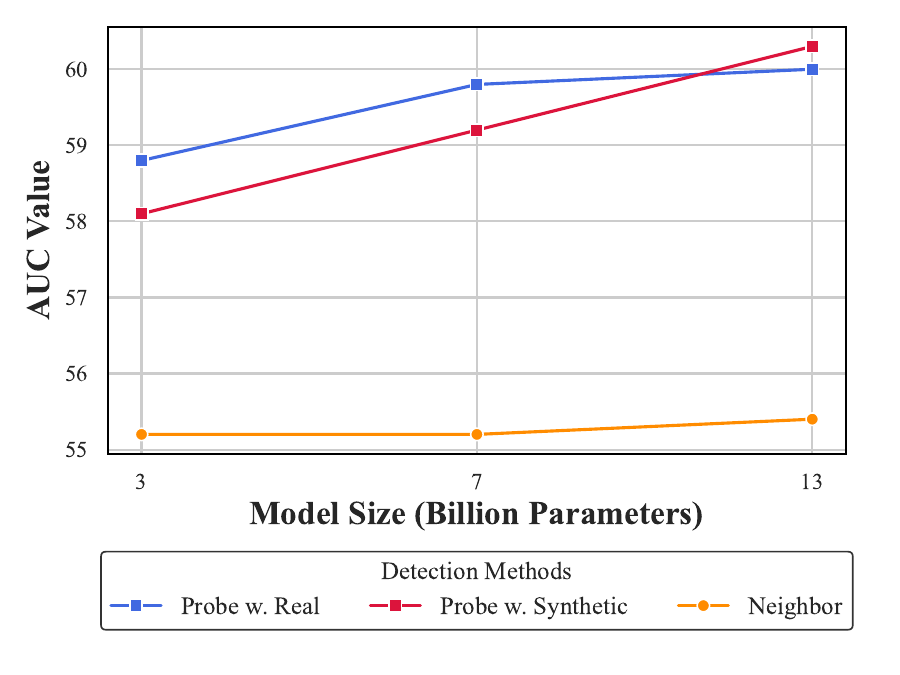}
    \caption{Comparison of AUC Values Across Different Model Sizes (best viewed in color).}
    \label{fig:model_size_scaling}
\end{figure}

\textbf{Model Size.} We evaluate our method and neighbor attack on ArxivMIA with different OpenLLaMA sizes (3B/7B/13B). As shown in \autoref{fig:model_size_scaling}, the AUC values of our method increase with the model size, while the change of neighbor attack is not significant. This result indicates that our method benefits from larger models.

\vspace{6pt}

\begin{figure}
    \centering
    \includegraphics[width=\linewidth]{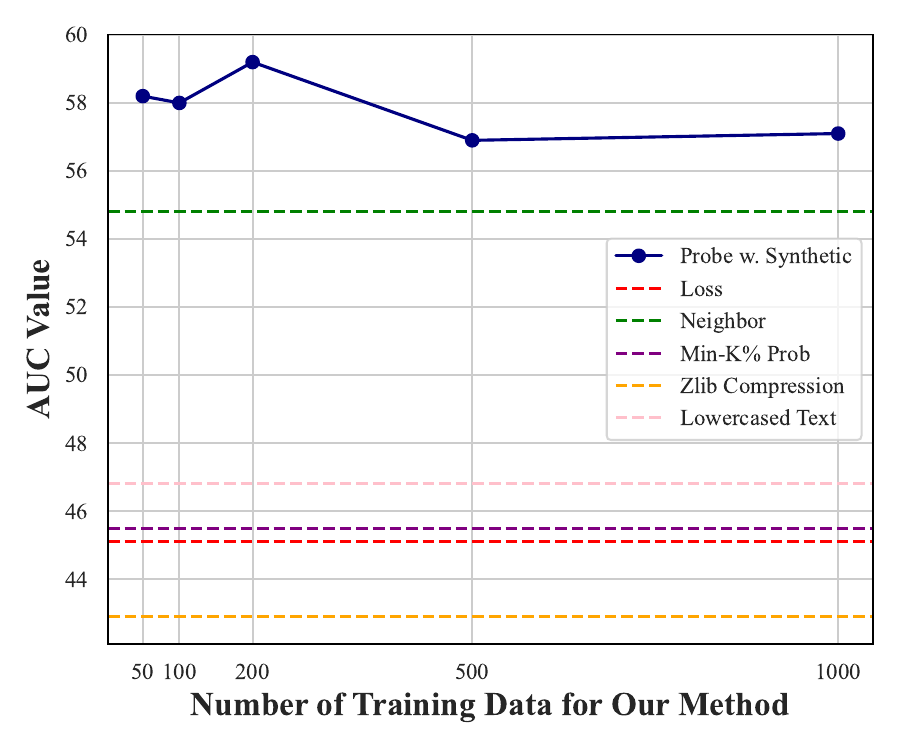}
    \caption{Comparison of AUC Values with Different Training Data Sizes (best viewed in color).}
    \label{fig:num_of_training_data}
\end{figure}

\textbf{Number of Training Data.} We also evaluate our method with different synthetic training data sizes (50, 100, 200, 500 and 1000). We conduct the comparison experiment on ArxivMIA with TinyLLaMA. As illustrated in \autoref{fig:num_of_training_data}, our method exhibits optimal performance with 200 training data samples. Increasing the number of training data beyond this point results in a slight decline in performance, yet it remains superior to various baselines. This indicates that our method is data efficient.

\subsection{Downstream Task Datasets Contamination Detection Challenge}

\begin{table}[h!]
    \centering
    \begin{tabular}{lcc}
        \toprule
        \textbf{Method} & \textbf{PMQA} & \textbf{CQA} \\
        \midrule
        \rowcolor[gray]{.93}\textbf{Reference-free Methods} & \multicolumn{2}{c}{} \\
        Loss Attack             & 48.0  & 49.9  \\
        Neighbor Attack         & 53.0  & 50.0  \\
        Min-K\% Prob            & 47.5  & 49.6  \\
        \midrule
        \rowcolor[gray]{.93}\textbf{Reference-based Methods} & \multicolumn{2}{c}{} \\
        Zlib Compression        & 46.1  & 48.8   \\
        Lowercased Text         & 50.7  & 49.2   \\
        Smaller Model           & 49.5  & 49.5   \\
        \midrule
        \rowcolor[gray]{.93}\textbf{Our Method} & \multicolumn{2}{c}{} \\
        Probe w. Synthetic Data & \textbf{54.0}  & \textbf{51.9}   \\
        \bottomrule
    \end{tabular}
    \caption{AUC values of various pre-training data detection methods on PubMedQA and CommonsenseQA in contamination detection challenge. PMQA denotes PubMedQA, CQA denotes CommonsenseQA. We highlight the best results in \textbf{bold}.}
    \label{tabs:result_of_contam_1.4b}
\end{table}

\begin{table}[h!]
    \centering
    \begin{tabular}{lc}
        \toprule
        \textbf{Method}                                      & \textbf{TinyLLaMA}   \\
        \midrule
        \rowcolor[gray]{.93}\textbf{Reference-free Methods}  & \multicolumn{1}{c}{} \\
        Loss Attack                                          & 55.7                 \\
        Neighbor Attack                                      & 48.8                 \\
        Min-K\% Prob                                         & 52.9                 \\
        \midrule
        \rowcolor[gray]{.93}\textbf{Reference-based Methods} & \multicolumn{1}{c}{} \\
        Zlib Compression                                     & 56.7                 \\
        Lowercased Text                                      & 49.8                 \\
        Smaller Model                                        & -                    \\
        \midrule
        \rowcolor[gray]{.93}\textbf{Our Method}              & \multicolumn{1}{c}{} \\
        Probe w. Real Data                                   & \textbf{74.3}        \\
        Probe w. Real Data (LoRA)                            & 62.7                 \\
        \bottomrule
    \end{tabular}
    \caption{AUC values of various methods on WikiMIA in contamination detection challenge. We highlight the best results in \textbf{bold}.}
    \label{tabs:result_of_peft}
\end{table}

To support the development of further work on detecting pretraining data contamination, \citet{oren2023proving} pre-trained a 1.4 billion parameter GPT-2 model \cite{radford2019language}, Contam-1.4b, with intentional downstream task datasets contamination \footnote{\url{https://github.com/tatsu-lab/test_set_contamination}}. We evaluate various detection methods on PubMedQA \cite{jin-etal-2019-pubmedqa} and CommonsenseQA \cite{talmor-etal-2019-commonsenseqa} from this challenge. PubMedQA and CommonsenseQA have different duplication counts (how often the dataset was injected into the pre-training data) with 1 and 2, and detection at this low duplication level is extremely difficult \cite{oren2023proving}.

\paragraph{Experimental Setup.} We sampled 1000 examples from the contaminated training data as member data for each task and then sampled 1000 examples from their standard dataset as non-member data. Similar to \autoref{sec:datasets}, we split each dataset into a validation set and a test set. The validation set will be used to select the best hyperparameters, and the test set for evaluation. For our method, we collected 200 synthetic training data for each task. For comparing to smaller model baseline setting, we choose Contam-Small (124M Params) pre-trained on the same dataset for Contam-1.4b.

\paragraph{Results.} The results are shown in \autoref{tabs:result_of_contam_1.4b}. We observe that our method outperforms other baselines, which demonstrate the effectiveness of our method. Nonetheless, we acknowledge that the overall detection efficacy is unsatisfactory at an extremely low duplication count (1 and 2), corroborating the findings of \citet{oren2023proving}.

\begin{table}[h!]
    \centering
    \begin{tabular}{ccc}
        \toprule
        Train/Eval        & \textbf{WikiMIA} & \textbf{ArxivMIA} \\
        \textbf{WikiMIA}  & 74.3             & 51.8              \\
        \textbf{ArxivMIA} & 73.2             & 57.1              \\
        \bottomrule
    \end{tabular}
    \caption{AUC values of TinyLLaMA across different domains.}
    \label{tabs:cross_domain}
\end{table}

\section{Conclusion}

In summary, this paper investigates the pre-training data detection problem in large language models. We propose a simple and effective approach that determines whether a target text has been included in a model's pre-training dataset by analyzing the internal activations using the probe technique. Additionally, we introduce a more challenging benchmark, ArxivMIA. The experiments demonstrate that our method outperforms all baselines across various benchmarks, achieving SOTA performance. We further analyze the impact of target model size and the number of training data on our method. Additionally, we validate the effectiveness of our approach through a downstream task datasets contamination detection challenge. Future work could extend our methods to larger model scales or apply them to multi-modal models.

\section*{Limitations}

\paragraph{Generalization.} One limitation of our study stems from the generalizability of the probe classifier, which necessitates domain-specific training data. This characteristic implies that the training data are not universally applicable across different domains/benchmarks. Consequently, to detect data from varied fields, it becomes imperative to collect distinct sets of training data for each domain.

\paragraph{Computational Resource Requirements.} While our method demonstrates superior performance, it necessitates a certain amount of computational resources due to the requirement to train both a proxy model and a probe classifier. We conduct experiments on WikiMIA with LoRA \cite{hu2022lora}, a representative Parameter-Efficient Fine-Tuning (PEFT) method. The results are presented in \autoref{tabs:result_of_peft}. We observed that the performance of LoRA is inferior to fine-tuning, but it remains competitive with other baselines. For our experimental settings, the training compute consumption is not that high since the member dataset is rather small. However, as the number of data scales, PEFT may be a better choice.

%%%%%
\section*{Acknowledgments}
This work is supported by the National Natural Science Foundation of China (Grant No. 62036004, 62376177) and Provincial Key Laboratory for Computer Information Processing Technology, Soochow University. This work is also supported by Collaborative Innovation Center of Novel Software Technology and Industrialization, the Priority Academic Program Development of Jiangsu Higher Education Institutions. We would also like to thank the anonymous reviewers for their insightful and valuable comments.
%%%%%

\bibliography{anthology,custom}

\appendix

\section{Data Synthesis with ChatGPT}
\label{sec:chatgpt_synthetic_data}

Given a target dataset \(D_0\), our goal is to utilize ChatGPT to generate a new, similar, domain-specific dataset \(D\). To achieve this, we employ a templated prompt to guide ChatGPT in generating data points that are stylistically and structurally similar to \(D_0\), yet unique in content. The prompt template used is shown in \autoref{tabs:data_generation_template}.

To initiate this process, we randomly select 5 examples from \(D_0\) and insert them into the prompt. This prompt is then provided to ChatGPT, which generates a specified number of new data points. By iterating through this procedure multiple rounds, we can get a dataset \(D\) that is similar to and within the same domain as \(D_0\).

\section{Prompt Template Selection}
\label{sec:prompt_template_selection}

We experiment with different prompt templates on ArxivMIA with TinyLLaMA to determine the optimal one for our task. The results of the ablation study on prompt templates are shown in \autoref{tabs:prompt_template_selection}. The average AUC value for using a template is 57.9, which is higher than the AUC value of 54.3 for not using a template. This demonstrates the importance of a prompt template.

\begin{table*}[h]
    \centering
    \begin{tabular}{|p{0.9\textwidth}|}
        \hline
        I am creating a dataset and need to generate data that is similar but not identical to the following examples. Here are 5 examples from my dataset:                                                                                                                                                                                  \\
        1. [Example 1]                                                                                                                                                                                                                                                                                                                       \\
        2. [Example 2]                                                                                                                                                                                                                                                                                                                       \\
        3. [Example 3]                                                                                                                                                                                                                                                                                                                       \\
        4. [Example 4]                                                                                                                                                                                                                                                                                                                       \\
        5. [Example 5]                                                                                                                                                                                                                                                                                                                       \\
        \\
        Please generate [Specified Number] new data points that are similar in style and structure to these examples but are unique in content. Format the responses as a numbered list, starting from 6 onwards. Each data point should start on a new line and be prefixed with its corresponding number followed by a period and a space. \\
        For example:                                                                                                                                                                                                                                                                                                                         \\
        6. [New Data Point 1]                                                                                                                                                                                                                                                                                                                \\
        7. [New Data Point 2]                                                                                                                                                                                                                                                                                                                \\
        ...                                                                                                                                                                                                                                                                                                                                  \\
        \hline
    \end{tabular}
    \caption{Data Generation Template.}
    \label{tabs:data_generation_template}
\end{table*}

\begin{table*}[ht]
    \centering
    \begin{tabular}{lr}
        \toprule
        \textbf{Prompt Template}                                                                   & \textbf{AUC} \\
        \midrule
        \textit{[SAMPLE]}                                                                          & 54.3         \\
        \midrule
        \textit{Here is a statement: [SAMPLE] Is the above statement correct? Answer:}             & 59.2         \\
        \midrule
        \textit{[SAMPLE] Is the above text real? Answer:}                                          & 57.1         \\
        \midrule
        \textit{Here is a text: [SAMPLE] Is the above text real? Answer:}                          & 58.3         \\
        \midrule
        \textit{Here is a text: [SAMPLE] Is the above text correct? Answer:}                       & 57.3         \\
        \midrule
        \textit{Consider the following statement: [SAMPLE]  Is the statement above true or false?} &              \\
        \textit{ Your answer:}                                                                     & 57.5         \\
        \midrule
        \textit{Consider the following text: [SAMPLE]  Is the text above true or false?}           &              \\
        \textit{ Your answer:}                                                                     & 59.1         \\
        \midrule
        \textit{[SAMPLE] Is the statement above true or false? Your answer:}                       & 57.8         \\
        \midrule
        \textit{[SAMPLE] Is this correct? Indicate 'Yes' or 'No': }                                & 57.0         \\
        \bottomrule
    \end{tabular}
    \caption{AUC values of different prompt templates on ArxivMIA with TinyLLaMA.}
    \label{tabs:prompt_template_selection}
\end{table*}

\end{document}